\documentclass[10pt,twocolumn,letterpaper]{article}

\usepackage{cvpr}
\usepackage{times}
\usepackage{epsfig}
\usepackage{graphicx}
\usepackage{amsmath}
\usepackage{amssymb}
\usepackage{subfig}
\usepackage{arydshln}
\usepackage{bm}

\usepackage{verbatim}
\usepackage{color}
\usepackage{lipsum}    
\usepackage{soul}

\usepackage[pagebackref=true,breaklinks=true,letterpaper=true,colorlinks,bookmarks=false]{hyperref}

\newcommand{\object}[1]{\texttt{#1}}
\newcommand{\predicate}[1]{\texttt{#1}}
\newcommand{\relationship}[3]{$<$\texttt{#1} - \texttt{#2} - \texttt{#3}$>$}

\newif\ifmain
\maintrue 
\newif\ifsupple
\suppletrue 

\cvprfinalcopy 

\def\cvprPaperID{792} 

\newcommand\blfootnote[1]{%
  \begingroup
  \renewcommand\thefootnote{}\footnote{#1}%
  \addtocounter{footnote}{-1}%
  \endgroup
}

\begin{document}

\title{Referring Relationships}

\author{
Ranjay Krishna$^{\dagger}$, Ines Chami$^{\dagger}$, Michael Bernstein, Li Fei-Fei\\
Stanford University\\
\small{\{ranjaykrishna, chami, msb, feifeili\}@cs.stanford.edu}
}

\maketitle
\ifmain
\begin{abstract}
Images are not simply sets of objects: each image represents a web of interconnected relationships. These relationships between entities carry semantic meaning and help a viewer differentiate between instances of an entity. For example, in an image of a soccer match, there may be multiple \object{person}s present, but each participates in different relationships: one is \predicate{kicking} the \object{ball}, and the other is \predicate{guarding} the \object{goal}. In this paper, we formulate the task of utilizing these ``referring relationships'' to disambiguate between entities of the same category. We introduce an iterative model that localizes the two entities in the referring relationship, conditioned on one another. We formulate the cyclic condition between the entities in a relationship by modelling predicates that connect the entities as shifts in attention from one entity to another. We demonstrate that our model can not only outperform existing approaches on three datasets --- CLEVR, VRD and Visual Genome --- but also that it produces visually meaningful predicate shifts, as an instance of interpretable neural networks. Finally, we show that by modelling predicates as attention shifts, we can even localize entities in the absence of their category, allowing our model to find completely unseen categories.
\end{abstract}

\section{Introduction}

\begin{figure}[t]
\begin{center}
   \includegraphics[width=\columnwidth]{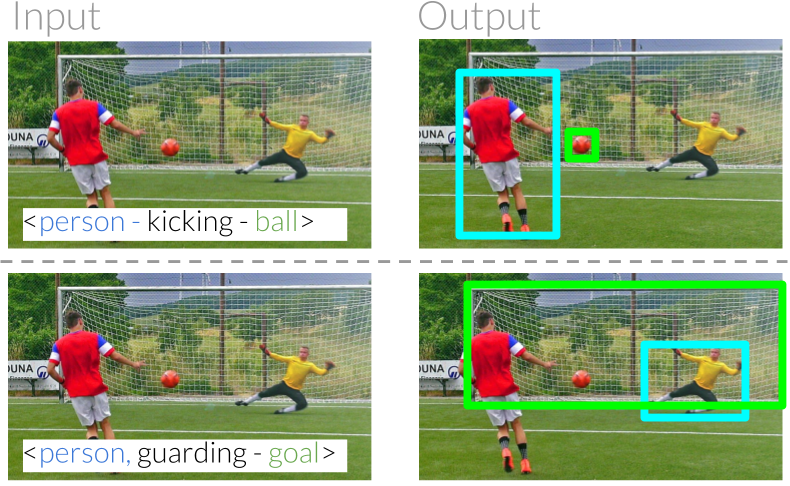}
\end{center}
   \caption{Referring relationships disambiguate between instances of the same category by using their relative relationships with other entities. Given the relationship \relationship{person}{kicking}{ball}, the task requires our model to correctly identify which \object{person} in the image is kicking the \object{ball} by understanding the predicate \predicate{kicking}.}
\label{fig:pull_figure}
\end{figure}

\blfootnote{$\dagger$ = equal contribution}
Referring expressions in everyday discourse help identify and locate entities\footnote{We use the term ``entities'' for what is commonly referred to as ``objects'' to differentiate from the term \object{object} in $<$subject-predicate-object$>$ relationships.} in our surroundings. For instance, we might point to the ``person kicking the ball'' to differentiate from the ``person guarding the goal'' (Figure~\ref{fig:pull_figure}). In both these examples, we disambiguate between the two \object{person}s by their respective relationships with other entities~\cite{lu2016visual}. While one \object{person} is \predicate{kicking} the \object{ball}, the other is \predicate{guarding} the \object{goal}. The eventual goal is to build computational models that can identify which entities others are referring to~\cite{shridhar2017grounding}.

To enable such interactions, we introduce referring relationships --- a task where, given a relationship, models should know which entities in a scene are being referred to by the relationship. Formally, the task expects an input image along with a relationship, which is of the form \relationship{subject}{predicate}{object}, and outputs localizations of both the \object{subject} and \object{object}. For example, we can express the above examples as \relationship{person}{kicking}{ball} and \relationship{person}{guarding}{goal} (Figure~\ref{fig:pull_figure}). Previous work has attempted to disambiguate entities of the same category in the context of referring expression comprehension~\cite{rohrbach2016grounding,mao2016generation,yu2016modeling,yu2016joint,hu2017modeling}. Their task expects a natural language input, such as ``a person guarding the goal'', resulting in evaluations that require both natural language as well as computer vision components. It can be challenging to pinpoint whether errors made by these models occur from either the language or the visual components. By interfacing with a structured relationship input, our task is a special case of referring expressions that alleviates the need to model language.

Referring relationships retain and refine the algorithmic challenges at the core of prior tasks. In the object localization literature, some entities such as \object{zebra} and \object{person} are highly discriminative and can be easily detected, while others such as \object{glass} and \object{ball} tend to be harder to localize~\cite{russakovsky2015imagenet}. These difficulties arise due to, for example, small size and non-discriminative composition. This difference in difficulty translates over to the referring relationships task. To tackle this challenge, we use the intuition that detecting one entity becomes easier if we know where the other one is. In other words, we can find the \object{ball} conditioned on the \object{person} who is \predicate{kicking} it and vice versa. We train this cyclic dependency by rolling out our model and iteratively passing messages between the subject and the object through an operator defined by the \predicate{predicate}. We describe this operator in more detail in Section~\ref{section:model}.

However, modelling this \predicate{predicate} operator is not straightforward, which leads us to our second challenge. Traditionally, previous visual relationship papers have learned an appearance-based model for each predicate~\cite{li2017vip,lu2016visual,plummer2016phrase}. Unfortunately, the drastic appearance variance of predicates, depending on the entities involved, makes learning predicate appearance models challenging. For example, the appearance for the predicate \predicate{carrying} can vary significantly between the following two relationships: \relationship{person}{carrying}{phone} and \relationship{truck}{carrying}{hay}. Instead, inspired by the moving spotlight theory in psychology~\cite{laberge1997shifting,sperling1995episodic}, we bypass this challenge by using predicates as a visual attention shift operation from one entity to the other. While one shift operation learns to move attention from the \object{subject} to the \object{object}, an inverse predicate shift similarly moves attention from the \object{object} back to the \object{subject}. Over multiple iterations, we operationalize these asymmetric attention shifts between the \object{subject} and the \object{object} as different types of message operations for each \predicate{predicate}~\cite{xu2017scene,gilmer2017neural}.

In summary, we introduce the task of referring relationships, whose structured relationship input allows us to evaluate how well we can unambiguously identify entities of the same category in an image. We evaluate our model\footnote{Our model was coded using Keras with a Tensorflow backend and is available at \url{https://github.com/StanfordVL/ReferringRelationships}.} on three vision datasets that contain visual relationships: CLEVR~\cite{johnson2016clevr}, VRD~\cite{lu2016visual} and Visual Genome~\cite{krishna2017visual}. $33\%$, $60.3\%$, and $61\%$ of relationships in these datasets refer to ambiguous entities, i.e.~entities that have multiple instances of the same category. We extend our model to perform attention saccades~\cite{torralba2006contextual} using relationships belonging to a scene graph~\cite{johnson2015image}. Finally, we demonstrate that in the absence of a \object{subject} or the \object{object}, our model can still disambiguate between entities while also localizing entities from new categories that it has never seen before.

\section{Related Work}

To properly situate the task of referring relationships, we explore the evolution of visual relationships as a representation. Next, we survey the inception of referring expression comprehension as a similar task, summarize how attention has been used in the deep learning literature, and survey other technical approaches that are similar to our approach.

There is a long history of vision papers moving beyond simple object detection and \textbf{modelling the context} around the entities~\cite{rabinovich2007objects,salakhutdinov2011learning} or even studying object co-occurrences~\cite{galleguillos2008object,ladicky2010graph,mensink2014costa} to improve classification and detection itself. Our task on referring relationships was motivated by such papers. Unlike these models, we utilize a formal definition for context in the form of a \textbf{visual relationship}. 

\begin{figure*}[t]
    \centering
    \includegraphics[width=\linewidth]{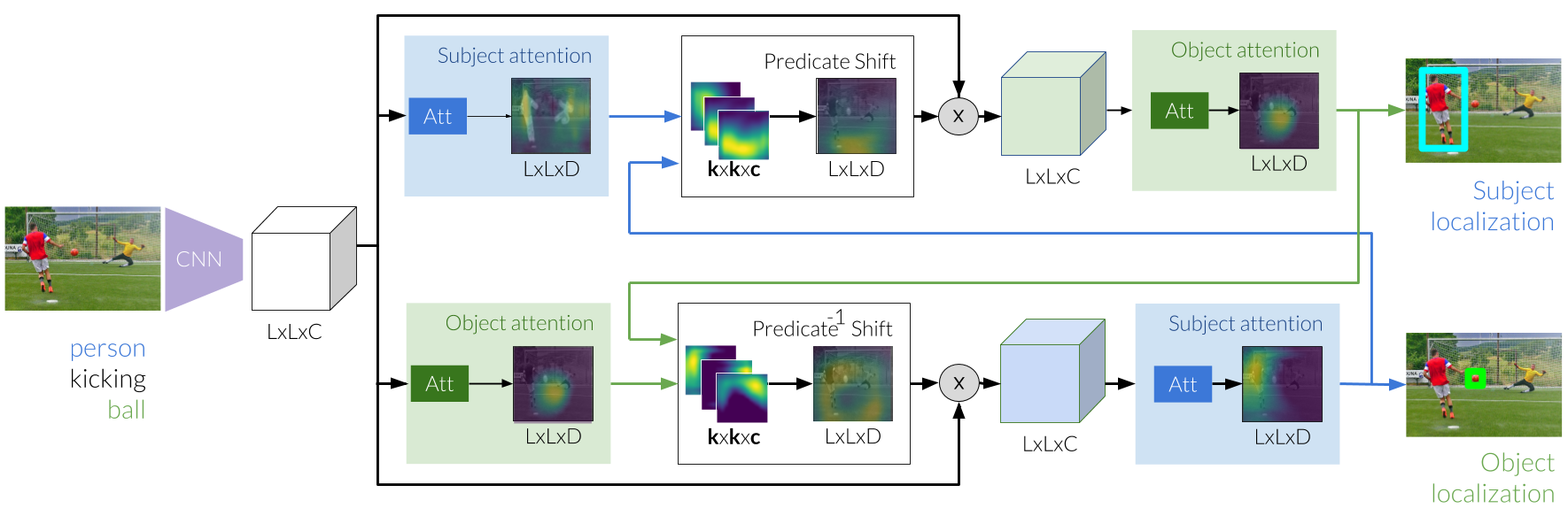}
    \caption{Referring relationships' inference pipeline begins by extracting image features, which are then used to generate an initial grounding of the \object{subject} and \object{object} independently. Next, these estimates are used to shift the attention using the \predicate{predicate} from the \object{subject} to where we expect the \object{object} to be. We modify the image features by focusing our attention to the shifted area when refining our new estimate of the \object{object}. Simultaneously, we learn an inverse shift from the initial \object{object} to the \object{subject}. We iteratively pass messages between the subject and object through the two predicate shift modules to finally localize the two entities.}
    \label{fig:model}
\end{figure*}

Pushing along this thread, visual relationships were initially limited to spatial relationships: \predicate{above}, \predicate{below}, \predicate{inside} and \predicate{around}~\cite{galleguillos2008object}. Relationships were then extended to include human interactions, such as \predicate{holding} and \predicate{carrying}~\cite{yao2010modeling}. Extending the definition further, the task of visual relationship detection was introduced along with a dataset of spatial, comparative, action and verb predicates~\cite{lu2016visual}. More recently, relationships were formalized as part of an explicit formal representation for images called scene graphs~\cite{johnson2015image,krishna2017visual}, along with a dataset of scene graphs called Visual Genome~\cite{krishna2017visual}. These scene graphs encode the entities in a scene as nodes in a graph that are connected together with directed edges representing their relative relationships. Scene graphs have shown to improve a number of computer vision tasks, including semantic image retrieval~\cite{schuster2015generating}, image captioning~\cite{anderson2016spice} and object detection~\cite{sadeghi2011recognition}. Newer work has extended models for relationship detection to use co-occurrence statistics~\cite{plummer2016phrase,santoro2017simple,xu2017scene} and have even formulated the problem in a reinforcement learning framework~\cite{liang2017deep}. These papers focused primarily on detecting visual relationships categorically --- they output relationships given an input image. In contrast, we focus on the inverse problem of localizing the entities that take part in an input relationship. We disambiguate entities in a query relationship from other entities of the same category in the image. Moreover, while all previous work has attempted to learn visual features of predicates, we propose that the visual appearances of predicates are too varied and can be more effectively learnt as an attention shift, conditioned on the entities in the relationship.

Such an inverse task of disambiguating between different regions in an image has been studied under the task of \textbf{referring expression comprehension}~\cite{mao2016generation}. This task uses an input language description to find the referred entities. This work has been motivated by human-robot interaction, where the robot would have to disambiguate which entities the human user is referring to~\cite{shridhar2017grounding}. Models for their task have been extended to include global image contrasts~\cite{yu2016modeling}, visual relationships~\cite{hu2017modeling} and reward-based reinforcement systems that encourage the generation of unique expressions for different image regions~\cite{yu2016modeling}. Unfortunately, all these models require the ability to process both natural language as well as visual constructs. This requirement makes it difficult to disentangle the mistakes as a result of poor language modelling or visual understanding. In an effort to ameliorate these limitations, we propose the referring relationships task --- simplifying referring expressions by replacing the language inputs with a structured relationship. We focus solely on the visual component of the model, avoiding confounding errors from language processing.

One key observations about predicates is their large variance in visual appearance~\cite{lu2016visual}. For example, consider these two relationships: \relationship{person}{carrying}{phone} and \relationship{truck}{carrying}{hay}. We use an insight from psychology~\cite{laberge1997shifting,sperling1995episodic}, specifically the moving spotlight theory, which suggests that visual attention can be modelled as a spotlight that can be conditioned on and directed towards specific targets. The use of attention has been explored to improve image captioning~\cite{xu2015show,bahdanau2014neural} and even stacked to improve question answering~\cite{johnson2017inferring,yang2016stacked}. In comparison, we model two discriminative \textbf{attention shifting} operations for each unique predicate, one conditioned on the \object{subject} to localize the \object{object} and an inverse predicate shift conditioned on the \object{object} to find the \object{subject}. Each predicate utilizes both the current estimate of the entities as well as image features to learn how to shift, allowing it to utilize both spatial and semantic features.

Our work also has similarities to \textbf{knowledge bases}, where predicates are often projections in a defined semantic space~\cite{bordes2013translating,dettmers2017convolutional,lin2015learning}. Such a method was recently used for visual relationship detection~\cite{zhang2017visual}. While these methods have seen success in knowledge base completion tasks, they have only led to a marginal gain for modelling visual relationships. However, unlike these methods, we do not model predicates as a projection in semantic space but as a shift in attention conditioned on an entity in a relationship. Our method can be thought of as a special case of deformable parts model~\cite{felzenszwalb2008discriminatively} with two deformable parts, one for each entity. Finally, our messaging passing algorithm can be thought of as a domain-specific specialized version to the message passing in graph convolution approximation methods~\cite{gilmer2017neural,kipf2016semi}.


\section{Referring relationships model}
\label{section:model}

Recall that our aim is to use the input referring relationship to disambiguate entities in an image by localizing the entities involved in the relationship. Formally, the input is an image $\mathbf{I}$ with a referring relationship, $R = $\relationship{S}{P}{O}, which are the \object{subject}, \predicate{predicate} and \object{object} categories, respectively. The model is expected to localize both the \object{subject} and the \object{object}.

\subsection{Problem formulation}
We begin by using a pre-trained convolutional neural network (CNN) to extract a $L\times L\times C$ dimensional feature map from the image $\boldsymbol{\mu} = \mathrm{CNN}(I)$. That is, for each image, we extract a 3-dimensional tensor of shape $L\times L\times C$, where $L$ is the spatial size of the feature map while $C$ is the number of feature channels. Our goal is to decide if each $L\times L$ image region belongs to the subject or object or neither. We can model this problem by representing the image by two binary random variables $\mathbf{X}, \mathbf{Y}$. For $i = 1 \ldots L\times L$, $X_{i} > \tau$ implies that the \object{subject} occupies the region $i$ and $Y_{i} > \tau$ implies that the \object{object} occupies that region, for some hyperparameter threshold $\tau$. 
We now define a graph $G = (\mathcal{V}_X \cup \mathcal{V}_Y, \mathcal{E})$, where $\mathcal{V}_X = \{x_{i}\}$, $\mathcal{V}_Y = \{y_{i}\}$ are the nodes of the graph represented by the image regions and $\mathcal{E} = (x_{i}, y_{j})$ represents an edge from every $x_i$ to $y_j$. Given the image and relationship, we want to assign $\mathbf{x}^*$ and $\mathbf{y}^*$ with $\mathbf{x}^*, \mathbf{y}^* = \arg \max_{\mathbf{x},\mathbf{y}} \mathrm{Pr}(\mathbf{X}=\mathbf{x}, \mathbf{Y}=\mathbf{y}|\boldsymbol{\mu}, R)$.

This optimization problem can be reduced to inference on a densely connected graph which can be very expensive. As shown in previous work~\cite{zheng2015conditional,krahenbuhl2011efficient}, dense graph inference can be approximated by mean field in Conditional Random Fields (CRF). Such papers allow fully differential inference assuming weighted gaussians as pairwise potentials~\cite{zheng2015conditional}. To achieve greater flexibility in a more principled training framework, we design a general model where the messaging passing during inference is a series of learnt convolutions. More specifically, we design our model with two types of modules: attention and predicate shift modules. While attention models attempt to locate a specific category in an image, the predicate shift modules learn to move attention from one entity to another.

\subsection{Symmetric stacked attention shifting (SSAS) model}
Before we specify our attention and shift operators, let's revisit the challenges in referring relationships to motivate our design decisions. The two challenges are (1) the difference in difficulty in object detection and (2) the drastic appearance variance of predicates. First, the difference in difficulty arises because some objects like \object{zebra} and \object{person} are highly discriminative and can be easily detected while others like \object{glass} and \object{ball} tend to be harder to localize. We can overcome this problem by conditioning the localization of one entity on the other. If we know where the \object{person} is, we should be able to estimate the location of the \object{ball} that they are \predicate{kicking}.

Second, predicates tend to vary in appearance depending on the objects involved in the relationship. To deal with the wide appearance variance of predicates, we move away from how previous work~\cite{lu2016visual} attempted to learn appearance features of  predicates and instead treat predicates as a mechanism for shifting the attention from one object to another. Relationships like \predicate{above} should learn to focus attention down from the \object{subject} when locating the \object{object}, and the predicate \predicate{left of} should focus the attention to the right of the \object{subject}. Inversely, once we locate the \object{object}, the model should use \predicate{left of} to focus attention to the left to confirm its initial estimate of the \object{subject}. Note that not all predicates are spatial, so we also ensure that we can model their visual appearances by conditioning the shifts on the image features as well.

\noindent \textbf{Attention modules.} 
With these design goals in mind, we formulate the attention module as an initial estimate of the \object{subject} and \object{object} localizations by approximating the maximizers $\mathbf{x^*}$, $\mathbf{y^*}$ with the soft attention $\mathrm{Att}(\cdot)$:
\begin{align}
\mathbf{\hat{x}}^0 = \mathrm{Att}(\boldsymbol{\mu}, S) &= \mathrm{ReLU}(\boldsymbol{\mu} \cdot \mathrm{Emb}(S))\\
\mathbf{\hat{y}}^0 = \mathrm{Att}(\boldsymbol{\mu}, O) &= \mathrm{ReLU}(\boldsymbol{\mu} \cdot \mathrm{Emb}(O))
\label{eq:Aopp}
\end{align}
where $Emb(\cdot)$ embeds the entity into a $C$ dimensional semantic space. Note that $\mathrm{ReLU(\cdot)}$ is the Rectified Linear Unit operator. $\mathbf{\hat{x}}^0$, $\mathbf{\hat{y}}^0$ denote the initial attention over the \object{subject} and \object{object}, which are not conditioned on the predicate at all and only use the entities.

\noindent \textbf{Predicate shift modules.} Inspired by the message passing protocol in CRF's \cite{zheng2015conditional}, we design a more general message passing function to transfer information between the two entities. Each message is passed from the \object{subject}'s estimate to localize the \object{object} and vice versa. In practice, we want the message passed from the \object{subject} to the \object{object} to be different from the one passed from the \object{object} back to the \object{subject}. So, we learn two asymmetric attention shifts, one that shifts the location from the \object{subject} to its estimate of where it thinks the \object{object} is and another one that does the inverse from the \object{object} to the \object{subject}. We denote these shift operations as $\mathrm{Sh}(\cdot)$ and $\mathrm{Sh}^{-1}(\cdot)$, respectively and define them as $n$ convolutions applied in series to the initial estimated assignments:
\begin{align}
\mathbf{\hat{x}}^0_{shift} &= \mathrm{Sh}^{-1}(\mathbf{\hat{y}}^0, P) = \bigcirc^n_l\mathrm{ReLU}(\mathbf{\hat{y}}^0 * F_{l}^{-1}(P))\\
\mathbf{\hat{y}}^0_{shift} &= \mathrm{Sh}(\mathbf{\hat{x}}^0, P) = \bigcirc^n_l\mathrm{ReLU}(\mathbf{\hat{x}}^0 * F_l(P)).
\end{align}
where the $\bigcirc^n_l$ implies that we perform the operation $n$ times, each parametrized by $F_l^{-1}(P)$ and $F_l(P)$ which correspond to learned convolution filters for the inverse predicate and the predicate operations respectively. The $*$ operator indicates a convolution with kernels $F_l^{-1}(P)$ and $F_l(P)$ of size $k_l=k$ with $c_l$ channels. We set $c_n=1$ for the last convolution to ensure that $\mathbf{\hat{x}}^0_{shift}$ and $\mathbf{\hat{y}}^0_{shift}$ have dimension $L\mathrm{x}L\mathrm{x}1$. While we do not enforce the two shift operators to be inverses of one another, for most predicates, we empirically find that $\mathrm{Sh}^{-1}(\cdot)$ in fact learns the inverse attention shift of $\mathrm{Sh}(\cdot)$. Note that we do not provide any supervision to our shifts and the model is tasked to learn these shifts to improve its entity localizations. The outputs of these two predicate shift operators is a new estimate attention mask over where the our model expects to find the \object{object}, $\mathbf{\hat{y}}^0_{shift}$, conditioned on its initial estimate of the \object{subject}, $\mathbf{\hat{x}}^0$ and vice versa from $\mathbf{\hat{y}}^0$ to $\mathbf{\hat{x}}^0_{shift}$.

Each predicate learns its own set of shift and inverse shift functions. And by allowing multiple channels $c_l$ for each set of kernels, our model can formulate shifts as a mixture. For example, \predicate{carrying} might want to focus on the top of the object when the relationship is \relationship{person}{carrying}{phone} while focusing towards the bottom when the relationship is \relationship{person}{carrying}{bag}.

Since we want every image region $X_{i}$ to pass a message to all other regions $Y_{j}$, we enforce that $n > L/k$, i.e.~we need a minimum of $L/k$ number of convolutions in series. We arrive at this restriction because the maximum spatial distance that a message needs to travel is $\sqrt{2}L$ and the furtherest image region it can send a message to in each iteration is $\sqrt{2}k$, where $L$ is the image feature size and $k$ is the kernel size of each predicate shift convolution.

\noindent \textbf{Running iterative inference.}
Once we have these estimates, we can modify our image features with using a element-wise multiplication across the $C$ channels in the feature map. We can then pass it back to the \object{subject} and \object{object} attention modules to update their locations:
\begin{align}
\mathbf{\hat{x}}^1 &= \mathrm{Att}(\mathbf{\hat{x}}^0_{shift} \times \boldsymbol{\mu}, S)\\
\mathbf{\hat{y}}^1 &= \mathrm{Att}(\mathbf{\hat{y}}^0_{shift} \times \boldsymbol{\mu}, O)
\end{align}
We can continuously update these locations, conditioned on one another. This amounts to running a maximum a posteriori inference on one entity while using the other entity's previous location. We finally output $\mathbf{\hat{x}}^t$ and $\mathbf{\hat{y}}^t$ where $t$ is a hyper-parameter that determines the number of iterations for which we run inference.

\noindent \textbf{Image Encoding.} We extract image features using an ImageNet pre-trained~\cite{russakovsky2015imagenet} ResNet50's~\cite{he2016deep} last activation layer of conv$4$ which outputs a $14\times 14\times 512$ dimensional representation and finetune the features. We find that our model performs best with predicate convolution filters with kernel size $5\times 5$ and $10$ channels.

\noindent \textbf{Training details.} We use RMSProp as our optimization function with an initial learning rate of $0.0001$ decaying by $30\%$ when the validation loss does not decrease for $3$ consecutive epochs. We train for a total of $30$ epochs and embed all of our objects and predicates in a $512$ dimensional space.


\section{Experiments}

\begin{table*}[t]
\begin{center}
\setlength{\tabcolsep}{4.7pt}
\begin{tabular}{ l c c c c c c    c c c c c c}
& \multicolumn{6}{ c }{Mean IoU $\uparrow$} & \multicolumn{6}{ c }{KL divergence $\downarrow$}\\
& \multicolumn{2}{ c }{CLEVR} & \multicolumn{2}{ c }{VRD} & \multicolumn{2}{ c }{Visual Genome} & \multicolumn{2}{ c }{CLEVR} & \multicolumn{2}{ c }{VRD} & \multicolumn{2}{ c }{Visual Genome}\\
 & S & O & S & O & S & O & S & O & S & O & S & O\\
\hline\hline
Co-occurence~\cite{galleguillos2008object} & 0.691 & 0.691 & 0.347 & 0.389 & 0.414& 0.490& 0.839 & 0.839& 2.598& 2.307& 1.501& 1.271\\
Spatial shift~\cite{laberge1997shifting} & 0.740& 0.740& 0.320& 0.371& 0.399& 0.469& 0.643& 0.643& 2.612& 2.318& 1.512& 1.293\\
VRD~\cite{lu2016visual,hu2017modeling} & 0.734 & 0.732 & 0.345 & 0.387 & 0.417& 0.480& 1.024 & 1.014& 2.492& 2.171& 1.483& 1.255\\
\hdashline
SSAS(iter1) & 0.742 & 0.748 & 0.358 & 0.398 & \textbf{0.426} & \textbf{0.491} & 0.623 & 0.640 & 1.936 & 1.710 & 1.483 & 1.235 \\
SSAS(iter2) & 0.777 & \textbf{0.779} & 0.365 & 0.404 & 0.422 & 0.487 & 0.597 & \textbf{0.595}& 1.783& \textbf{1.549}& 1.458 & 1.212\\
SSAS(iter3) & \textbf{0.778} & 0.778 & \textbf{0.369} & \textbf{0.410}& 0.421 & 0.482 & \textbf{0.595} & 0.596 & \textbf{1.741}& 1.576& \textbf{1.457} & \textbf{1.205}\\
\hline
\end{tabular}
\caption{Results for referring relationships on CLEVR~\cite{johnson2016clevr}, VRD~\cite{lu2016visual} and Visual Genome~\cite{krishna2017visual}. We report Mean IoU and KL divergence for the subject and object localizations individually.}
\label{tab:results}
\end{center}
\end{table*}

We start our experiments by evaluating our model's performance on referring relationships across three datasets, where each dataset provides a unique set of characteristics that complement our experiments. Next, we evaluate how to improve our model in the absence of one of the entities in the input referring relationship. Finally, we conclude by demonstrating how our model can be modularized and used to perform attention saccades through a scene graph.

\subsection{Datasets and Baselines}
\noindent \textbf{CLEVR}. CLEVR is a synthetic dataset generated from scene graphs~\cite{johnson2016clevr}, where the relationships between objects are limited to $4$ spatial predicates (\predicate{left}, \predicate{right}, \predicate{front}, \predicate{behind}) and $48$ distinct entity categories.
With over $5M$ relationships where $30\%$ are ambiguous, along with the ease of localizing object categories, this dataset also allows us to explicitly test the effects of our predicate attention shifts without confounding errors from poor image features or noise in real world datasets. 

\noindent \textbf{VRD}. Visual relationship detection (VRD) is the most widely benchmarked dataset for relationship detection in real world images~\cite{lu2016visual}. It consists of $100$ object and $70$ predicate categories in $5k$ images, with $60\%$ ambiguous relationships out of a total of $38k$. With a few examples per object and predicate category, this dataset allows us to evaluate how our model performs when starved for data.

\noindent \textbf{Visual Genome}. Visual Genome is the largest dataset for visual relationships in real images that is publicly available~\cite{krishna2017visual}. It contains $100k$ images with over $2.3M$ relationship instances. We use version $1.4$, which focuses on the $100$ most common objects with the $70$ most common predicate categories. Our experiments on Visual Genome represent a large scale evaluation of our method where $61\%$ of relationships refer to ambiguous entities.

\noindent \textbf{Evaluation Metrics.} Recall that the output of our model is localizing the subject and the object of the referring relationship. To evaluate how our model performs, we report the Mean Intersection over Union (IoU), a common metric used in localizing salient parts of an image~\cite{bylinskii2015saliency,bylinskii2016should}. This metric measures the average intersection over union between the predicted image regions to those in the ground truth bounding boxes. Next, we report the KL-divergence, which measures the dissimilarity between the two saliency maps and heavily penalizes false positives.

\noindent \textbf{Baseline models.} We create three competitive baseline models inspired by related work in entity co-occurrence~\cite{galleguillos2008object}, spatial attention shifts~\cite{laberge1997shifting} and visual relationship detection~\cite{lu2016visual}. The first model tests how much we can leverage only the entities' \textbf{co-occurrence}, without using the predicate. This model simply embeds the \object{subject} and the \object{object} and combines them to collectively attend over the image features. The next baseline embeds the entities along with the predicate using a series of dense layers, similar to the vision component in relationship embeddings used in visual relationship detection (\textbf{VRD})~\cite{lu2016visual,hu2017modeling}. This model has access to the entire relationship when finding the two entities. Finally, the third baseline replaces our learnt predicate shifts with a \textbf{spatial shift} that we statistically learn for each predicate in the dataset (see supplementary for details). This final model tests whether our model utilizes both semantic information from images and not just the spatial information from the entities to make predictions.

\begin{figure}[t]
    \centering
    \subfloat[]{{\includegraphics[width=\columnwidth]{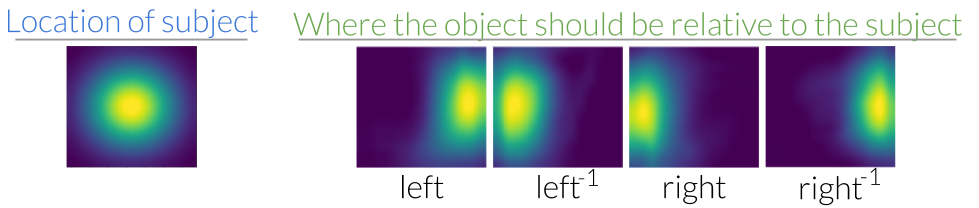} }}
    \qquad
  	\subfloat[]{{\includegraphics[width=0.85\linewidth]{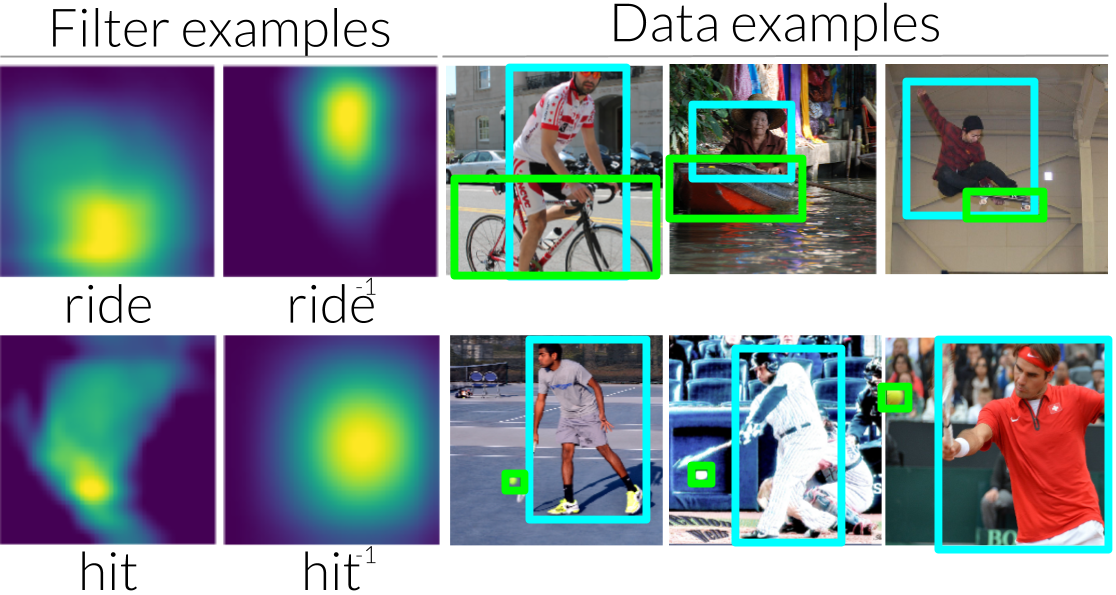} }}
    \caption{(a) Relative to a subject in the middle of an image, the predicate \predicate{left} will shift the attention to the right when using the relationship \relationship{subject}{left of}{object} to find the object. Inversely, when using the \object{object} to find the \object{subject}, the inverse predicate \predicate{left} will shift the attention to the left. We visualize all $70$ VRD, $6$ CLEVR and $70$ Visual Genome predicate and inverse predicate shifts in our supplementary material. (b) We also show that these shifts are intuitive when looking at the dataset that was used to learn them. For example, we find that \predicate{ride} usually corresponds to an \object{object} below the \object{subject}.}
    \label{fig:filters}
\end{figure}

\begin{figure*}[t]
    \centering
    \subfloat[]{{\includegraphics[width=0.8\linewidth]{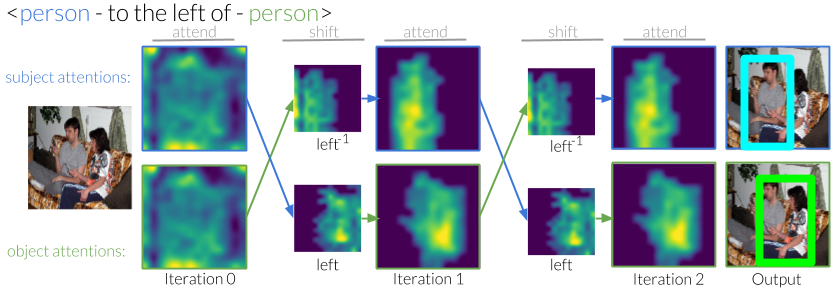} }}
    \qquad
    \subfloat[]{{\includegraphics[width=0.66\columnwidth]{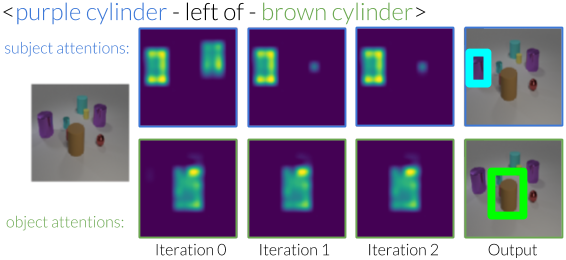} }}
    \subfloat[]{{\includegraphics[width=0.66\columnwidth]{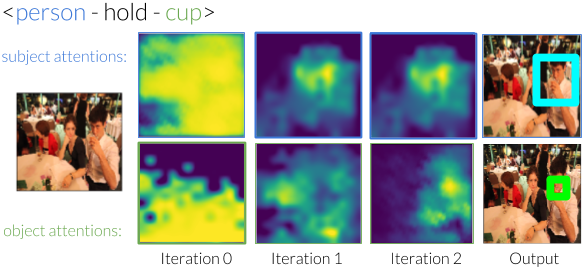} }}
    \subfloat[]{{\includegraphics[width=0.66\columnwidth]{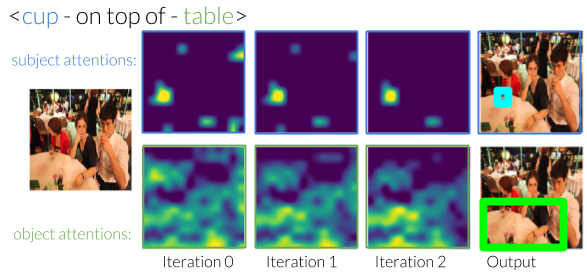} }}
    \qquad
    \subfloat[]{{\includegraphics[width=0.66\columnwidth]{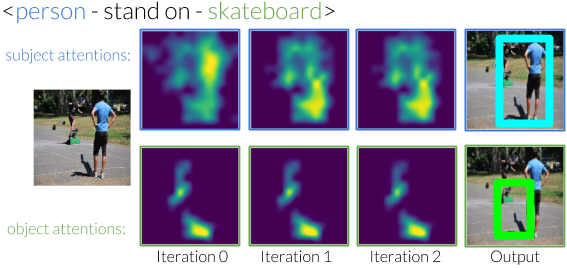} }}
    \subfloat[]{{\includegraphics[width=0.66\columnwidth]{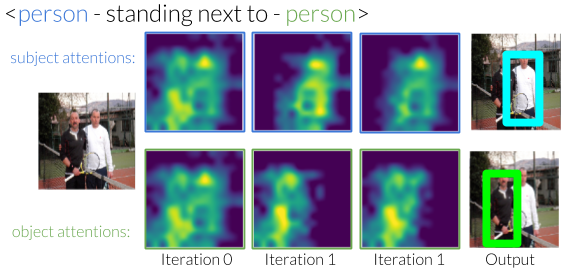} }}
    \subfloat[]{{\includegraphics[width=0.66\columnwidth]{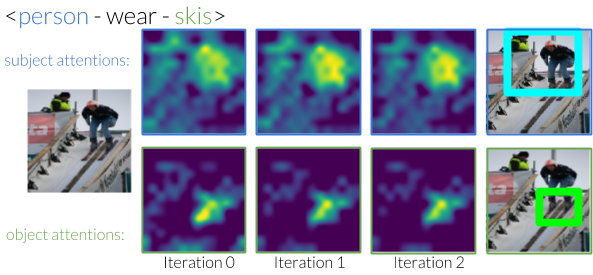} }}
    \qquad
    \caption{Example visualizations of how attention shifts across multiple iterations from the CLEVR and Visual Genome datasets. On the first iteration the model receives information only about the entities that it is trying to find and hence attempts to localize all instances of those categories. In later iterations, we see that the predicate shifts the attention, allowing our model to disambiguate between different instances of the same category.}
    \label{fig:example}
\end{figure*}

\subsection{Results}
\noindent \textbf{Quantitative results.}  
Across all the datasets, we find that the \textbf{co-occurrence} model is unable to disambiguate between instances of the same category and only performs well when there is only one instance of that category in an image. The \textbf{spatial shift} model does better than the other baselines on CLEVR, where the predicates are spatial and worse on the real world datasets, implying that it is insufficient to model predicates only as spatial shifts. Surprisingly, when evaluating on the CLEVR dataset, we find that \textbf{VRD} model does not properly utilize the predicate and leads to marginal gains over the \textbf{co-occurrence} models. In comparison, we find that our \textbf{SSAS} variants perform better across all metrics. We gain over a $0.32$ Mean IoU on CLEVR. This gain however, is smaller on Visual Genome and VRD as these datasets are noisy and incomplete, penalizing our model for making predictions that are not annotated in the datasets. KL, which only penalizes false predictions highlights that our models are more precise than our baselines. Across the different ablations of SSAS, we notice that having more iterations is better; but the performance saturates after $3$ iterations because the predicate shifts and the inverse predicate shifts learn near inverse operations of one another.

\noindent \textbf{Interpreting our results.} We can interpret the predicate shifts by synthetically initializing the subject to be at the center of an image, as shown in Figure~\ref{fig:filters}(a). When applying the \predicate{left} predicate shift, we see that the model has learnt to focus its attention to the right, expecting to find the \object{object} to the right of the \object{subject}. Similarly, the inverse predicate shift learns to do nearly the opposite by focusing attention in the other direction. When visualizing these shifts next to the dataset examples in Visual Genome, we see that the shifts represent the biases that exist in the dataset (Figure~\ref{fig:filters}(b)). For example, since most entities that can be \predicate{ridden} are below the \object{subject}, the shifts learn to focus attention down to find the \object{object} and up to find the \object{subject}. We also find that that our model learns to encode dataset bias in these shifts. Since the perspective of most images in the training set for \predicate{hit} are of people playing tennis or baseball facing left, our model also captures this bias by learning that \predicate{hit} should focus attention to the bottom left to find the entity being hit.

Figure~\ref{fig:example} shows numerous examples of how our model shifts attention over multiple iterations. We see that generally across all our test cases the subject and object attention modules learn to use the image features to localize all instances initially on iteration $0$. For example, in Figure~\ref{fig:example}(a), all the regions that contain \object{person} are initially activated. But after the predicate and the inverse predicate shifts, we see that the model learns to move the attention in opposite directions for the predicate \predicate{left}. In the second iteration, both the people are uniquely localized in the image.  Figure~\ref{fig:example}(b) clearly shows that we can easily locate all instances of \object{purple metal cylinders} in the image since it is easy to detect entities in CLEVR. Our model learns to identify which \object{purple metal cylinders} we are actually referring to on successive iterations while suppressing the other instance. 

\begin{figure*}[t]
    \centering
  \includegraphics[width=\linewidth]{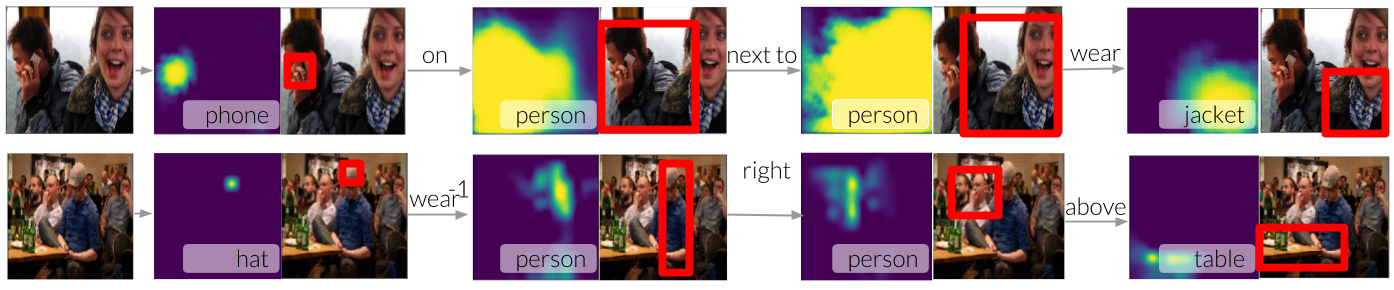}
    \caption{We can decompose our model into its attention and shift modules and stack them to attend over the nodes of a scene graph. Here we demonstrate how our model can be used to start at one node (\object{phone}) and traverse a scene graph using the relationships to connect the nodes and localize all the entities in the phrase $<$\texttt{phone on the person next to another person wearing a jacket}$>$. A second examples attends over the entities in $<$\texttt{hat worn by person to the right of another person above the table}$>$.}
    \label{fig:temporal_heatmap}
\end{figure*}

In Figure~\ref{fig:example}(c), even though both the subject and object have multiple instances of \object{person} and \object{cup}, we can disambiguate which \object{person} is actually \predicate{holding} the \object{cup}. For the same image in Figure~\ref{fig:example}(d), our model is able to distinguish the \object{cup} being \predicate{held} in the previous referring relationship from the one that is \predicate{on top of} the \object{table}. In cases where a referring relationship is not unique, like the example in Figure~\ref{fig:example}(e), we manage to find all instances that satisfy the relationship we care about. Here, we return both \object{person}s \predicate{riding} the \object{skateboard}s. Having learnt from the dataset, that most relationships with \predicate{stand next to} annotate the \object{subject} to the left of the \object{object}, our model emulates this behaviour in Figure~\ref{fig:example}(f). However, our model does make a fair share of mistakes - for example, in Figure~\ref{fig:example}(g), it finds both the \object{person}s and isn't able to distinguish which one is wearing the \object{skis}.

\begin{table}
\begin{center}
\setlength{\tabcolsep}{4.75pt}
\begin{tabular}{ l c c c c }
& No subject & No object & \multicolumn{2}{ c }{Only predicate}\\
& S-IoU & O-IoU & S-IoU & O--IoU \\
\hline\hline
VRD~\cite{lu2016visual} & 0.208 & 0.008 & 0.024 & 0.026 \\
SSAS (iter 1) & 0.331 & 0.359 & 0.332 & 0.361 \\
SSAS (iter 2) & 0.333 & 0.360 & \textbf{0.334} & 0.361\\
SSAS (iter 3) & \textbf{0.335} & \textbf{0.363} & \textbf{0.334} & \textbf{0.365} \\
\hline
\end{tabular}
\caption{Referring relationships results in the absence of the entities under three test conditions: \textbf{no subject} where the input is \relationship{\_\_\_}{predicate}{object}, \textbf{no object} where the input is \relationship{subject}{predicate}{\_\_\_} and \textbf{only predicate} where the input is \relationship{\_\_\_}{predicate}{\_\_\_}}
\vspace{3mm}
\label{tab:discovery}
\end{center}
\end{table}

\subsection{Localizing unseen categories}
Now that we have evaluated our model, one natural question to ask is how important is it for the model to receive both the entities of the relationship as input? Can it localize the \object{person} from Figure~\ref{fig:pull_figure} if we only use \relationship{\_\_\_}{kicking}{ball} as input? Or can we localize both the subject and the object with only \relationship{\_\_\_}{kicking}{\_\_\_}? We are also interested in taking this task a step further and studying whether we can localize categories that we have never seen before. Previous work has shown that we can localize \textbf{seen categories} in novel relationship combinations~\cite{lu2016visual} but we want to know if it is possible to localize \textbf{unseen categories}. 

We remove all instances of categories like \object{pants}, \object{hydrant}, etc. that are not in ImageNet ($\mathrm{CNN(\cdot)}$ was pre-trained on ImageNet) from our training set and attempt to localize these novel categories using their relationships. We do not make any changes to our model but alter the training script to randomly (we use a drop rate of $0.3$) mask out the \object{subject} or \object{object} or both in the referring relationships during each iteration. The model learns to attend over general object categories when the entities are masked out. We find that we can in fact localize these missing entities, even if they are from unseen categories. We report results for this experiment on the VRD dataset in Table~\ref{tab:discovery}.

\subsection{Attention saccades through a scene graph}
A ramification of our model design results in its modularity --- the attention and shift modules expect inputs and produce outputs that are image features of shape $L\times L\times C$. We can decompose these modules and stack them like Lego blocks, allowing us to perform more complicated tasks. One particularly interesting extension to referring relationships is attention saccades~\cite{torralba2006contextual}. Instead of using a single relationship as input, we can extend our model to take an entire scene graph as input. Figure~\ref{fig:temporal_heatmap} demonstrates how we can iterate between the attention and shift modules to traverse a scene graph. We can start from the \object{phone} and can localize the \object{jacket} worn by the ``woman on the right of the man using the phone''. A scene graph traversal can be evaluated by decomposing the graph into a series of relationships. We do not quantitatively evaluate these saccades here, as its evaluations are already captured by the referring relationships in the graph.

\section{Conclusion}
We introduced the task of referring relationships, where our model utilizes visual relationships to disambiguate between instances of the same category. Our model learns to iteratively use predicates as an attention shift between the two entities in a relationship. It updates its belief of where the \object{subject} and \object{object} are by conditioning its predictions on the previous location estimate of the \object{object} and \object{subject}, respectively. We show improvements on CLEVR, VRD and Visual Genome datasets. We also demonstrate that our model produces interpretable predicate shifts, allowing us to verify that the model is in fact learning to shift attention. We even showcase how our model can be used to localize completely unseen categories by relying on partial referring relationships and how it can be extended to perform attention saccades on scene graphs. Improvements in referring relationships could pave the way for vision algorithms to detect unseen entities and learn to grow its understanding of the visual world.

\noindent \textbf{Acknowledgements.} Toyota Research Center (TRI) provided funds to assist the authors with their research but this article solely reflects the opinions and conclusions of its authors and not TRI or any other Toyota entity. We thank  John Emmons, Justin Johnson and Yuke Zhu for their helpful comments.

\fi

\ifsupple

\section{Supplementary material}
In the supplementary material, we include more detailed results of our task for every entity and predicate category, allowing us to diagnose which entities or predicates are difficult to model. We also include the learnt predicate and the inverse predicate shifts for all $70$, $4$ and $70$ predicates we modeled in VRD~\cite{lu2016visual}, CLEVR~\cite{johnson2016clevr} and Visual Genome~\cite{krishna2017visual}. Furthermore, we explain our baseline models in more detail here.

\subsection*{Co-occurrence and VRD baseline models}
Given that the closest task to referring relationships is referring expression comprehension~\cite{mao2016generation}, we draw inspiration from this literature when designing our baselines. A frequent approach used by most models for this task involve semantically mapping language expressions to their corresponding image regions~\cite{rohrbach2016grounding,mao2016generation,yu2016modeling}. In other words, they map the image features extracted from a CNN close to the language expression features extracted from a Long Short Term Memory (LSTM). Our baseline models (\textbf{co-occurrence} and \textbf{VRD}) draws inspiration from this line of work and maps relationships to a semantic feature space and maps them close to the image regions to which they refer to using our attention module.

The difference from the two baseline models is determined by how we embed the relationships to that semantic space. In the case of \textbf{co-occurrence}, we are only interested in studying how well we can model relationship without the predicate and rely simply on co-occurrence statistics. So, we first embed the \object{subject} and the \object{object}, concatenate their representations and pass them through a dense layer followed by a RELU non-linearity to allow the two embeddings to interact. For the \textbf{VRD} baseline, we embed the entire relationship similar to prior work~\cite{lu2016visual} by embeddings all three components of the relationship, concatenating their representation and passing them through a dense and non-linear layer. 

Unlike our model, which attends over the subject and object in succession, these models are jointly aware of the entire relationship or at least about the other entity when attending over the image features. Also embedding the predicate and attending over the image with this embedding asks these baselines to model predicates as visual. But predicates such as \predicate{above} or \predicate{below} are not visually significant and can only be modelled as a relative shift from one entity to another. We show through our experiments that such baselines are not able to perform as well as our model nor are interpretable.

\subsection*{Spatial shift baseline model}
Instead of learning the attention shifts for each predicate, we assume (incorrectly) that all predicates are simply spatial shifts and model each predicate as a shift function. We learn the shift statistically from the relative locations of the two entities of the relationship. We visualize these statistically calculated shifts in Figures~\ref{fig:vrd_gt_predicates},~\ref{fig:clevr_gt_predicates} and~\ref{fig:visualgenome_gt_predicates}. We normalize the shifts so visualize the heatmaps. They don't show the actual values of how much each predicate shifts attention but only shows the direction of the shift. We see the as expected \predicate{left} push attention to the right, etc. This baseline uses our attention modules to find the subject and object and uses these precalculated shifts to move attention around. We only need to train the attention module, which is equivalent to training our SSAS model with zero iterations. During evaluation, we use these statistical spatial shifts to move attention. 

This baseline is useful in two ways. First, it demonstrates that it is important to model predicates as both spatial as well as semantic. Second, it allows us to compare the learnt predicate shifts with these calculated ones to verify that our SSAS models are in fact learning spatial shifts as well.

\subsection{Learnt predicate shifts}
While \predicate{above} and \predicate{below} are spatial predicates, others like \predicate{hit} or \predicate{sleep on} are both spatial as well as semantic. \predicate{hit} usually refers to entities around the \object{subject} and are usually \object{ball}s. Similarly, \predicate{sleep on} usually refers to something below the \object{subject} and typically a \object{bed} or \object{couch}. We show the learnt predicate shifts of all the predicates in the three datasets in Figures~\ref{fig:vrd_predicates},~\ref{fig:clevr_predicates} and~\ref{fig:visualgenome_predicates}.

As expected most relationships that are spatial are interpretable. In Figure~\ref{fig:vrd_predicates}, \predicate{above} moves attention below while its inverse moves it up. \predicate{hit} focuses on the right bottom, emulating the dataset bias of right handed people hitting tennis or baseball. In Figure~\ref{fig:visualgenome_predicates}, \predicate{wearing} shifts attention all over the body of the \object{subject} focusing mainly on \object{shirt}s, \object{pant}s and \object{glass}es. \predicate{By} splits the attention both to the left and to the right to find what the \object{subject} is next to. Some predicates, like \predicate{attached to} are harder to interpret as they depend on both the semantic as well as spatial shifts. While our model uses the image features to learn these shifts, our current spatial shift visualization does not create an interpretable predicate shift.

\subsection{Predicate analysis}
One of the benefits of referring relationships is its structured representation of the visual world, allowing us to study which entities and predicates are hard to model. In this section we report the Mean IoU of our model on all the predicate categories for the three datasets in Tables~\ref{tab:vrd_per_predicate} and~\ref{tab:visualgenome_per_predicate}. Note that we don't report the results for CLEVR here since all the $4$ spatial predicates are equally represented in the dataset and perform equally across all categories. 

Across most predicates we find that the \object{object} localization is much harder than the \object{subject}'s. This occurs because most \object{object}s tend to be smaller objects which are better localized by first attending over the \object{subject} first. We also see that size is an important factor in detection as predicates like \predicate{carry} and \predicate{use} usually have a larger \object{subject} and a smaller \object{object} and we find that the IoU for the \object{subject} is much higher than that of the \object{object}. We also see that when entities are partially occluded, for example \relationship{subject}{drive}{object}, the \object{object} IoU is much higher than the occluded \object{subject}.

\subsection{Object analysis}
We run a similar analyze of the performance of our model across all the entity categories and report Mean IoU results in Tables~\ref{tab:vrd_per_object} and~\ref{tab:visualgenome_per_object}. Note that we don't report the results for CLEVR here since all the entities perform equally across all categories.

We find that the Mean IoU for all entities in Visual Genome are higher than the ones in VRD, implying that more data for each of these categories helps the model learn to attend over the right image regions. In Figure~\ref{tab:visualgenome_per_object}, we find that with the predicate shifts, we can detect smaller objects, like \object{face}, \object{ear}, \object{bowl}, \object{eye},  a lot better. Some entities like \object{shelves} and \object{light} don't perform well on the dataset because not all the shelves or light sources are annotated in the dataset, causing the model's correct predictions to be penalized. Surprisingly, the model has a hard time finding \object{bag}s, perhaps because it learns that bags are often found being worn or carried by people in the training set but the test set contains bags that are on the ground or resting against other entities.

\begin{table*}[t]
\centering
\begin{tabular}{l c c | l c c | l c c}
Predicate & S-IoU & O-IoU & Predicate & S-IoU & O-IoU & Predicate & S-IoU & O-IoU \\
\hline\hline
on & 0.2904 & 0.5482 & wear & 0.4189 & 0.2830 & has & 0.4490 & 0.2339\\
next to & 0.3338 & 0.3867 & outside of & - & 0.7778 & sit next to & 0.3158 & 0.3152\\
stand next to & 0.4429 & 0.4436 & park next & 0.4012 & 0.5426 & sleep on & 0.3543 & 0.5429\\
above & 0.5653 & 0.4525 & behind & 0.3055 & 0.4770 & stand behind & 0.5748 & 0.4424\\
sit behind & 0.5854 & 0.9111 & park behind & 0.8545 & 0.5050 & in the front of & 0.3644 & 0.4009\\
under & 0.4639 & 0.5188 & stand under & 0.2304 & 0.3622 & sit under & 0.2716 & 0.3158\\
near & 0.2964 & 0.3642 & rest on & 0.4283 & 0.4603 & walk & 0.5814 & 0.6667\\
walk past & 0.6000 & 0.8571 & in & 0.3073 & 0.4339 & below & 0.4272 & 0.5337\\
beside & 0.2939 & 0.3870 & follow & 0.4249 & 0.5367 & over & 0.5403 & 0.5055\\
hold & 0.3867 & 0.1535 & by & 0.2705 & 0.4423 & beneath & 0.4888 & 0.5282\\
with & 0.3522 & 0.2823 & on the top of & 0.2896 & 0.4416 & on the left of & 0.2290 & 0.3272\\
on the right of & 0.2864 & 0.3338 & sit on & 0.4281 & 0.4271 & ride & 0.4513 & 0.4936\\
carry & 0.3334 & 0.1744 & look & 0.3344 & 0.2951 & stand on & 0.3854 & 0.7179\\
use & 0.4726 & 0.1160 & at & 0.2995 & 0.5185 & attach to & 0.4193 & 0.6047\\
cover & 0.3349 & 0.4364 & touch & 0.3426 & 0.4461 & watch & 0.3022 & 0.3982\\
against & 0.1364 & 0.6898 & inside & 0.1779 & 0.4751 & adjacent to & 0.7539 & 0.6492\\
across & 0.4460 & 0.5010 & contain & 0.3174 & 0.2443 & drive & 0.1168 & 0.6528\\
drive on & 0.7723 & 0.8269 & taller than & 0.4431 & 0.4423 & eat & 0.4726 & -\\
park on & 0.4639 & 0.7347 & lying on & 0.3457 & 0.6335 & pull & 0.4737 & 0.3362\\
talk & 0.7453 & 0.1767 & lean on & 0.5046 & 0.5127 & fly & 0.4517 & 0.2156\\
face & 0.3219 & 0.5598 & play with & 0.5735 & 0.2647 & \\
\end{tabular}
\caption{Mean IoU results for referring relationships per predicate in the VRD~\cite{lu2016visual} dataset.}
\label{tab:vrd_per_predicate}
\end{table*}

\begin{table*}[t]
\centering
\begin{tabular}{l c c | l c c | l c c}
Entity & S-IoU & O-IoU & Entity & S-IoU & O-IoU & Entity & S-IoU & O-IoU \\
\hline\hline
person & 0.3909 & 0.4191 & sky & 0.7651 & 0.7602 & building & 0.3635 & 0.4707\\
truck & 0.4477 & 0.5754 & bus & 0.5864 & 0.6578 & table & 0.4693 & 0.5664\\
shirt & 0.3495 & 0.3231 & chair & 0.2103 & 0.2448 & car & 0.3293 & 0.3764\\
train & 0.5213 & 0.5688 & glasses & 0.1682 & 0.2324 & tree & 0.3106 & 0.3398\\
boat & 0.2832 & 0.4775 & hat & 0.2368 & 0.2606 & trees & 0.4637 & 0.5840\\
grass & 0.5393 & 0.5474 & pants & 0.3612 & 0.3161 & road & 0.6776 & 0.6812\\
motorcycle & 0.5031 & 0.5291 & jacket & 0.3288 & 0.3316 & monitor & 0.3130 & 0.3404\\
wheel & 0.3348 & 0.2370 & umbrella & 0.2670 & 0.3426 & plate & 0.2011 & 0.2899\\
bike & 0.4091 & 0.3479 & clock & 0.2273 & 0.2193 & bag & 0.0951 & 0.0915\\
shoe & - & 0.1143 & laptop & 0.3319 & 0.3178 & desk & 0.5790 & 0.5945\\
cabinet & 0.1700 & 0.1845 & counter & 0.3477 & 0.4249 & bench & 0.3671 & 0.4308\\
shoes & 0.2944 & 0.2879 & tower & 0.4315 & 0.5556 & bottle & 0.1052 & 0.0809\\
helmet & 0.2834 & 0.2533 & stove & 0.2242 & 0.2941 & lamp & 0.1467 & 0.1692\\
coat & 0.2897 & 0.3203 & bed & 0.6702 & 0.6631 & dog & 0.3619 & 0.3510\\
mountain & 0.3915 & 0.4803 & horse & 0.5253 & 0.5527 & plane & 0.3193 & 0.6164\\
roof & 0.2859 & 0.2709 & skateboard & 0.4013 & 0.3694 & traffic light & 0.1067 & 0.0238\\
bush & 0.2328 & 0.2312 & phone & 0.0514 & 0.0671 & airplane & 0.5333 & 0.6694\\
sofa & 0.4597 & 0.5251 & cup & 0.1423 & 0.1030 & sink & 0.2592 & 0.2119\\
shelf & 0.0583 & 0.1278 & box & 0.0442 & 0.0996 & van & 0.2144 & 0.3710\\
hand & 0.1124 & 0.0413 & shorts & 0.2423 & 0.2547 & post & 0.0941 & 0.0971\\
jeans & 0.2449 & 0.3517 & cat & 0.3629 & 0.3238 & sunglasses & 0.3065 & 0.1535\\
bowl & 0.2226 & 0.0494 & computer & 0.2196 & 0.1676 & pillow & 0.1321 & 0.1797\\
pizza & 0.3882 & 0.3359 & basket & 0.1330 & 0.0751 & elephant & 0.1761 & 0.4534\\
kite & 0.2463 & 0.1843 & sand & 0.9597 & 0.7765 & keyboard & 0.2713 & 0.2421\\
plant & 0.1793 & 0.1275 & can & 0.1605 & 0.2452 & vase & 0.1575 & 0.2536\\
refrigerator & 0.1489 & 0.1949 & cart & 0.5619 & 0.5016 & skis & 0.1761 & 0.3398\\
pot & 0.1117 & 0.0450 & surfboard & 0.2676 & 0.2227 & paper & 0.1525 & 0.0296\\
mouse & 0.1164 & 0.1029 & trash can & 0.0324 & 0.0692 & cone & 0.1767 & 0.1813\\
camera & 0.0124 & 0.1183 & ball & 0.0595 & 0.0556 & bear & 0.3661 & 0.3441\\
giraffe & 0.5695 & 0.5949 & tie & 0.1129 & 0.1221 & luggage & 0.4560 & 0.5042\\
faucet & 0.1704 & 0.0565 & hydrant & 0.4108 & 0.5458 & snowboard & 0.2798 & 0.1804\\
oven & 0.4968 & 0.3169 & engine & 0.2016 & 0.1450 & watch & - & 0.0233\\
face & 0.0873 & 0.1798 & street & 0.6986 & 0.7291 & ramp & 0.2341 & 0.4972\\
\end{tabular}
\caption{Mean IoU results for referring relationships per entity category in the VRD~\cite{lu2016visual} dataset.}
\label{tab:vrd_per_object}
\end{table*}

\begin{table*}[t]
\centering
\begin{tabular}{l c c | l c c | l c c}
Predicate & S-IoU & O-IoU & Predicate & S-IoU & O-IoU & Predicate & S-IoU & O-IoU \\
\hline\hline
wearing a & 0.5208 & 0.3946 & made of & 0.4430 & 0.3389 & on front of & 0.2215 & 0.6592\\
with a & 0.4370 & 0.1098 & WEARING & 0.5125 & 0.3856 & above & 0.4642 & 0.4879\\
carrying & 0.4559 & 0.1555 & has an & 0.6672 & 0.0836 & covering & 0.6003 & 0.6558\\
and & 0.4192 & 0.1644 & wears & 0.5044 & 0.3542 & around & 0.4524 & 0.5527\\
with & 0.4923 & 0.3324 & laying on & 0.4557 & 0.6832 & inside & 0.2695 & 0.6084\\
attached to & 0.2627 & 0.4524 & at & 0.4473 & 0.5085 & on a & 0.3471 & 0.4978\\
of a & 0.2968 & 0.5857 & hanging on & 0.3166 & 0.4830 & near & 0.3931 & 0.4935\\
OF & 0.3320 & 0.6058 & sitting on & 0.4301 & 0.5331 & of & 0.3215 & 0.6172\\
next to & 0.3620 & 0.4949 & riding & 0.4959 & 0.4981 & under & 0.4276 & 0.5446\\
over & 0.3719 & 0.5039 & behind & 0.3798 & 0.5849 & sitting in & 0.4025 & 0.4852\\
ON & 0.3394 & 0.5508 & eating & 0.5277 & 0.4358 & to & 0.2768 & 0.5984\\
in a & 0.3580 & 0.4629 & has & 0.6183 & 0.3341 & parked on & 0.3851 & 0.5559\\
covered in & 0.5683 & 0.4607 & holding & 0.4716 & 0.3225 & for & 0.2892 & 0.4015\\
playing & 0.5863 & 0.5625 & against & 0.3765 & 0.5524 & by & 0.3368 & 0.4593\\
from & 0.2940 & 0.5188 & has a & 0.5841 & 0.3016 & standing on & 0.4715 & 0.6338\\
on side of & 0.2453 & 0.5505 & in & 0.3574 & 0.5320 & wearing & 0.4466 & 0.1613\\
watching & 0.3033 & 0.4851 & walking on & 0.4062 & 0.5990 & beside & 0.3592 & 0.5406\\
below & 0.4370 & 0.5168 & IN & 0.4005 & 0.5802 & mounted on & 0.3054 & 0.5426\\
have & 0.5750 & 0.2201 & are on & 0.3510 & 0.6001 & are in & 0.4185 & 0.6917\\
in front of & 0.3963 & 0.5210 & looking at & 0.4503 & 0.4787 & belonging to & 0.3250 & 0.6243\\
on top of & 0.3803 & 0.5735 & holds & 0.5194 & 0.3834 & inside of & 0.2398 & 0.3430\\
along & 0.3647 & 0.5030 & hanging from & 0.2508 & 0.2905 & standing in & 0.4748 & 0.6173\\
says & 0.1200 & - & painted on & 0.2632 & 0.6049 & between & 0.4090 & 0.4987\\
\end{tabular}
\caption{Mean IoU results for referring relationships per predicate in Visual Genome~\cite{krishna2017visual}.}
\label{tab:visualgenome_per_predicate}
\end{table*}

\begin{table*}[t]
\centering
\begin{tabular}{l c c | l c c | l c c}
Entity & S-IoU & O-IoU & Entity & S-IoU & O-IoU & Entity & S-IoU & O-IoU \\
\hline\hline
giraffe & 0.6361 & 0.6468 & bowl & 0.2602 & 0.3144 & food & 0.4410 & 0.4512\\
face & 0.3762 & 0.4020 & people & 0.3210 & 0.3492 & shirt & 0.3950 & 0.3774\\
bench & 0.4204 & 0.5398 & light & 0.1561 & 0.1574 & head & 0.3918 & 0.4283\\
zebra & 0.6152 & 0.6127 & cow & 0.5079 & 0.5867 & sign & 0.2390 & 0.3593\\
motorcycle & 0.5093 & 0.5648 & floor & 0.4870 & 0.5673 & hat & 0.3891 & 0.3270\\
sheep & 0.4988 & 0.4735 & truck & 0.4420 & 0.6199 & water & 0.4134 & 0.5766\\
chair & 0.2987 & 0.3421 & field & 0.6578 & 0.7331 & door & 0.2338 & 0.2906\\
pizza & 0.6415 & 0.4513 & tree & 0.3440 & 0.4077 & car & 0.3467 & 0.5168\\
leg & 0.3136 & 0.3383 & bag & 0.1794 & 0.1512 & fence & 0.4272 & 0.5216\\
sidewalk & 0.3623 & 0.4631 & girl & 0.5544 & 0.5707 & leaves & 0.2408 & 0.2376\\
jacket & 0.4363 & 0.3913 & windows & 0.2832 & 0.2672 & road & 0.5047 & 0.5897\\
glass & 0.2339 & 0.2186 & bed & 0.4867 & 0.6171 & sand & 0.4527 & 0.6028\\
trees & 0.4799 & 0.4973 & player & 0.6028 & 0.6511 & helmet & 0.3699 & 0.3809\\
man & 0.5355 & 0.5971 & grass & 0.4306 & 0.5724 & cake & 0.4235 & 0.4622\\
bear & 0.6530 & 0.6794 & hand & 0.2257 & 0.2279 & cloud & 0.4259 & 0.3843\\
street & 0.4765 & 0.5590 & ground & 0.6269 & 0.6302 & airplane & 0.6671 & 0.7176\\
mirror & 0.2132 & 0.3290 & clock & 0.4131 & 0.4533 & plate & 0.4529 & 0.5599\\
ear & 0.3029 & 0.2670 & hair & 0.3790 & 0.4054 & window & 0.2284 & 0.2473\\
boy & 0.5793 & 0.6432 & clouds & 0.4570 & 0.4644 & handle & 0.0671 & 0.1023\\
counter & 0.3018 & 0.4660 & glasses & 0.3164 & 0.3113 & pants & 0.4308 & 0.3939\\
eye & 0.2933 & 0.2427 & pole & 0.2374 & 0.2408 & line & 0.2265 & 0.2230\\
wall & 0.3599 & 0.4230 & animal & 0.4067 & 0.5630 & shadow & 0.3007 & 0.3013\\
train & 0.6389 & 0.6494 & bike & 0.5360 & 0.5238 & boat & 0.3467 & 0.4689\\
horse & 0.5631 & 0.5964 & tail & 0.3167 & 0.3189 & nose & 0.2959 & 0.2667\\
beach & 0.6542 & 0.6755 & snow & 0.5374 & 0.5755 & elephant & 0.6877 & 0.6409\\
bottle & 0.2039 & 0.1981 & surfboard & 0.3388 & 0.3861 & cat & 0.6501 & 0.6796\\
skateboard & 0.4036 & 0.4373 & shorts & 0.4454 & 0.3732 & woman & 0.5019 & 0.5392\\
bird & 0.4211 & 0.5768 & sky & 0.6741 & 0.7468 & shelf & 0.1316 & 0.1928\\
tracks & 0.3826 & 0.4737 & kite & 0.4496 & 0.3150 & umbrella & 0.3590 & 0.4102\\
guy & 0.5813 & 0.6980 & building & 0.4169 & 0.5366 & dog & 0.5649 & 0.6532\\
background & 0.5510 & 0.5531 & table & 0.3601 & 0.5719 & child & 0.4880 & 0.4252\\
lady & 0.5255 & 0.6257 & plane & 0.6689 & 0.6667 & desk & 0.3536 & 0.4990\\
bus & 0.6549 & 0.7362 & wheel & 0.2778 & 0.2744 & arm & 0.2747 & 0.2918\\
\end{tabular}
\caption{Mean IoU results for referring relationships per entity category in Visual Genome~\cite{krishna2017visual}.}
\label{tab:visualgenome_per_object}
\end{table*}

\subsection{CLEVR annotations}
\begin{figure}[t]
    \centering
    \includegraphics[width=0.8\columnwidth]{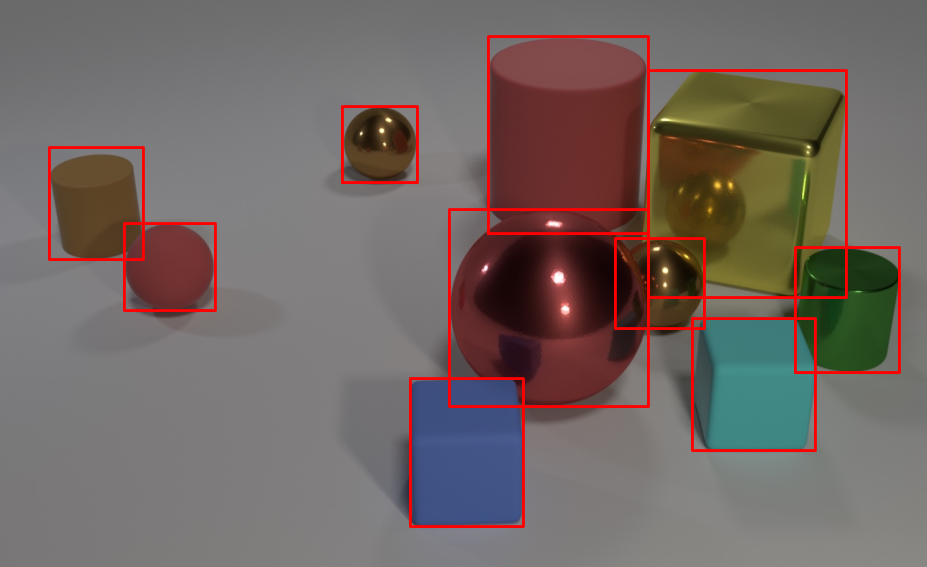}
    \caption{Example bounding box annotations we added to the Clevr dataset.}
    \label{fig:clevr_bbox}
\end{figure}

The CLEVR dataset is annotated with objects in 3D space~\cite{johnson2016clevr}. To use the dataset in the same manner as VRD~\cite{lu2016visual} and  VisualGenome~\cite{krishna2017visual}, we converted all the 3D entity locations into 2D bounding boxes, with respect to the viewing perspective of every image. We will release the conversion code as well as the bounding box annotations that we added to CLEVR. Figure~\ref{fig:clevr_bbox} showcases an example image annotated with our bounding boxes.

\begin{figure*}[t]
    \centering
    \includegraphics[width=0.8\linewidth]{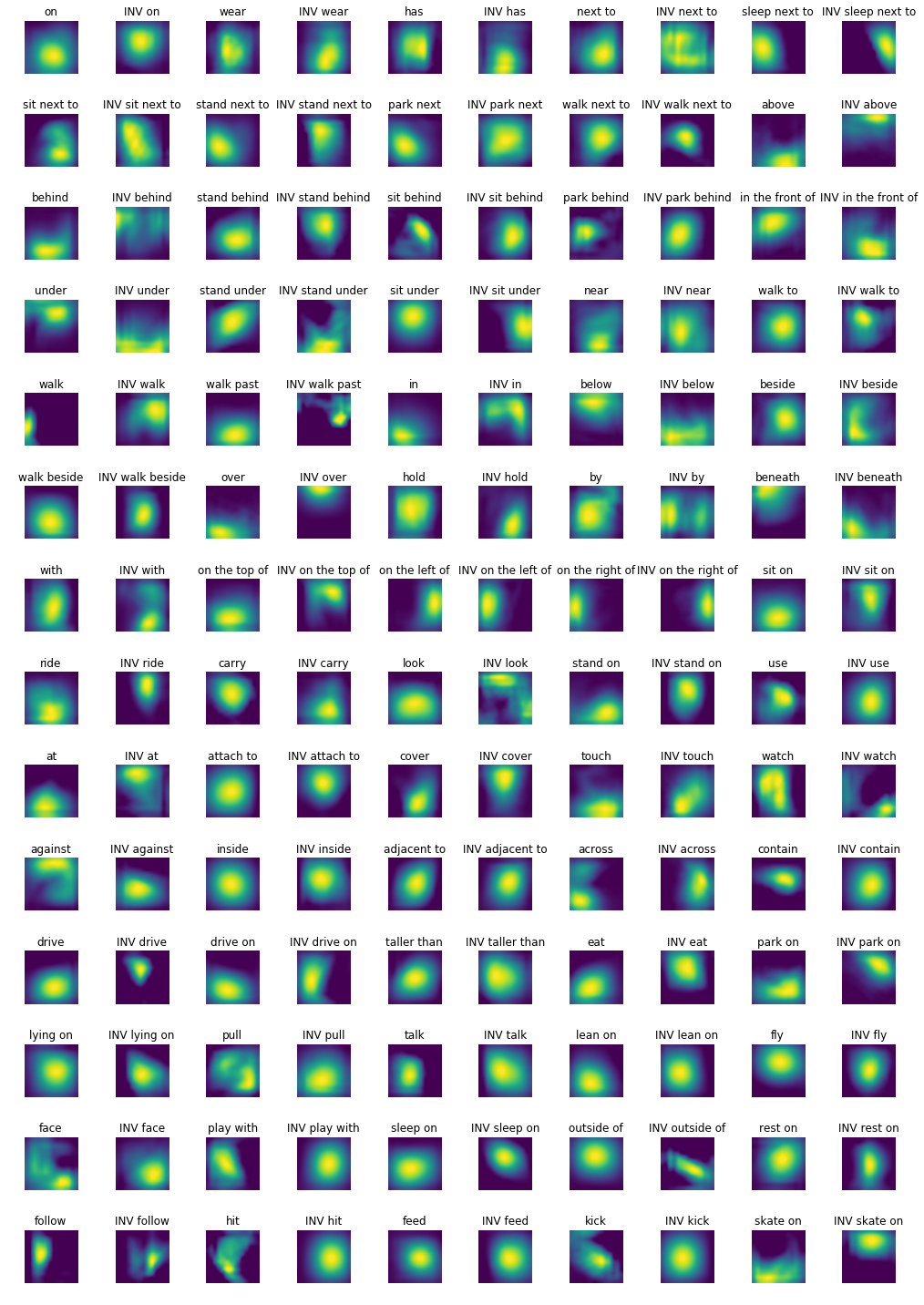}
    \caption{Learnt predicate shifts from the VRD dataset.}
    \label{fig:vrd_predicates}
\end{figure*}

\begin{figure*}[t]
    \centering
    \includegraphics[width=0.8\linewidth]{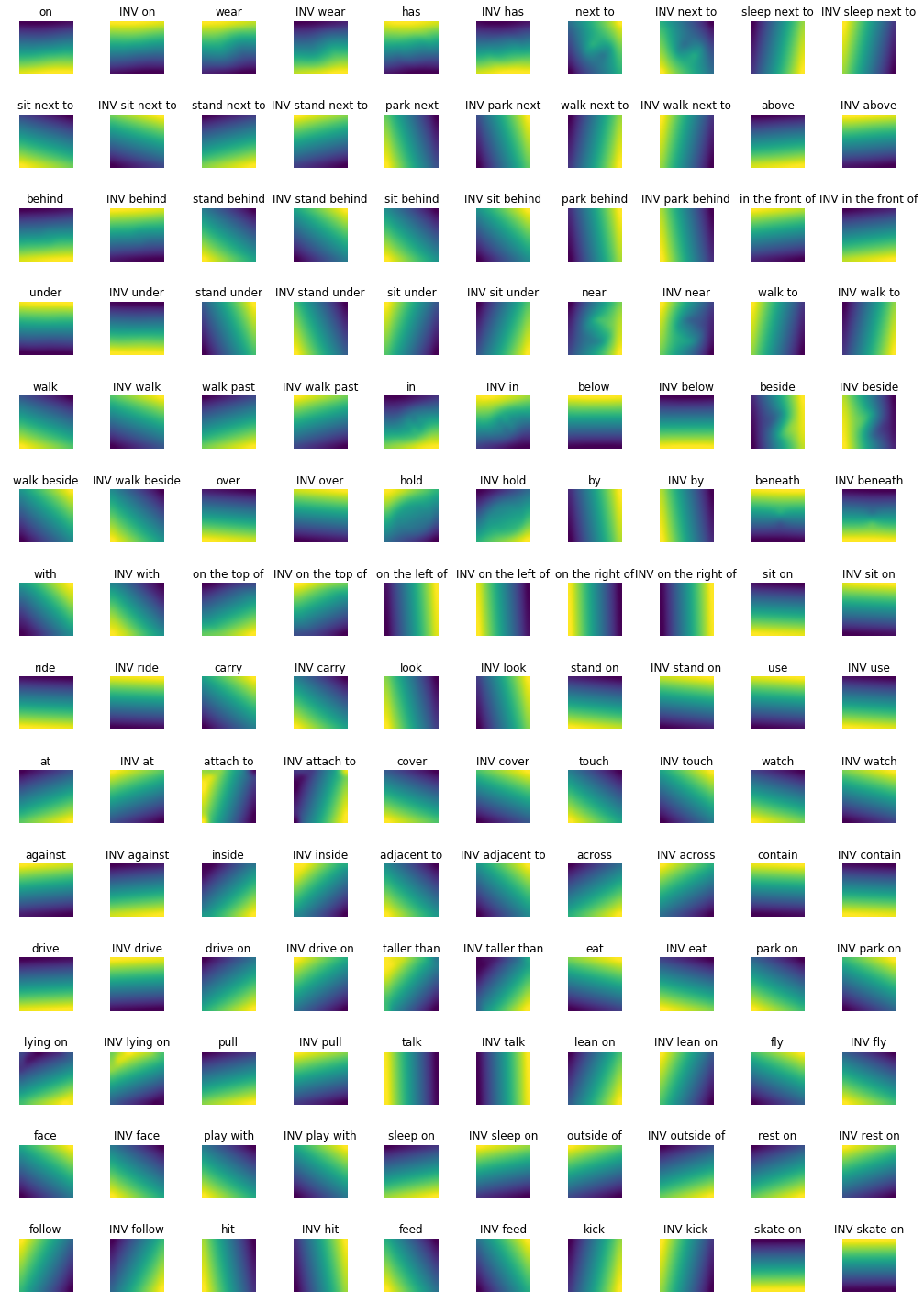}
    \caption{Spatial shifts calculated from the VRD dataset. These shifts were used for the \textbf{spatial shift} baseline model.}
    \label{fig:vrd_gt_predicates}
\end{figure*}

\begin{figure*}[t]
    \centering
    \includegraphics[width=0.9\columnwidth]{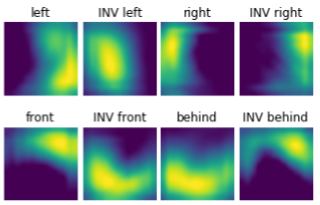}
    \caption{Learnt predicate shifts from the CLEVR dataset.}
    \label{fig:clevr_predicates}
\end{figure*}

\begin{figure*}[t]
    \centering
    \includegraphics[width=0.9\columnwidth]{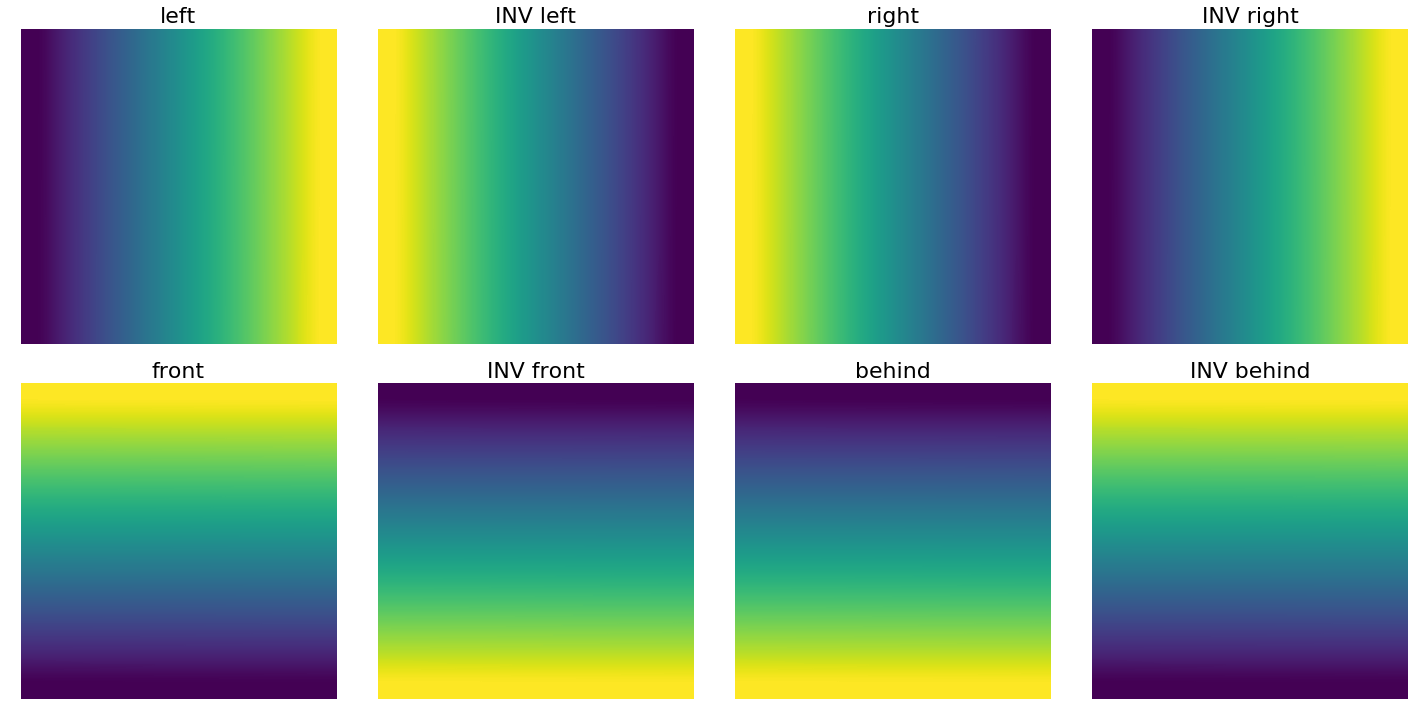}
    \caption{Spatial shifts calculated from the CLEVR dataset. These shifts were used for the \textbf{spatial shift} baseline model.}
    \label{fig:clevr_gt_predicates}
\end{figure*}

\begin{figure*}[t]
    \centering
    \includegraphics[width=0.8\linewidth]{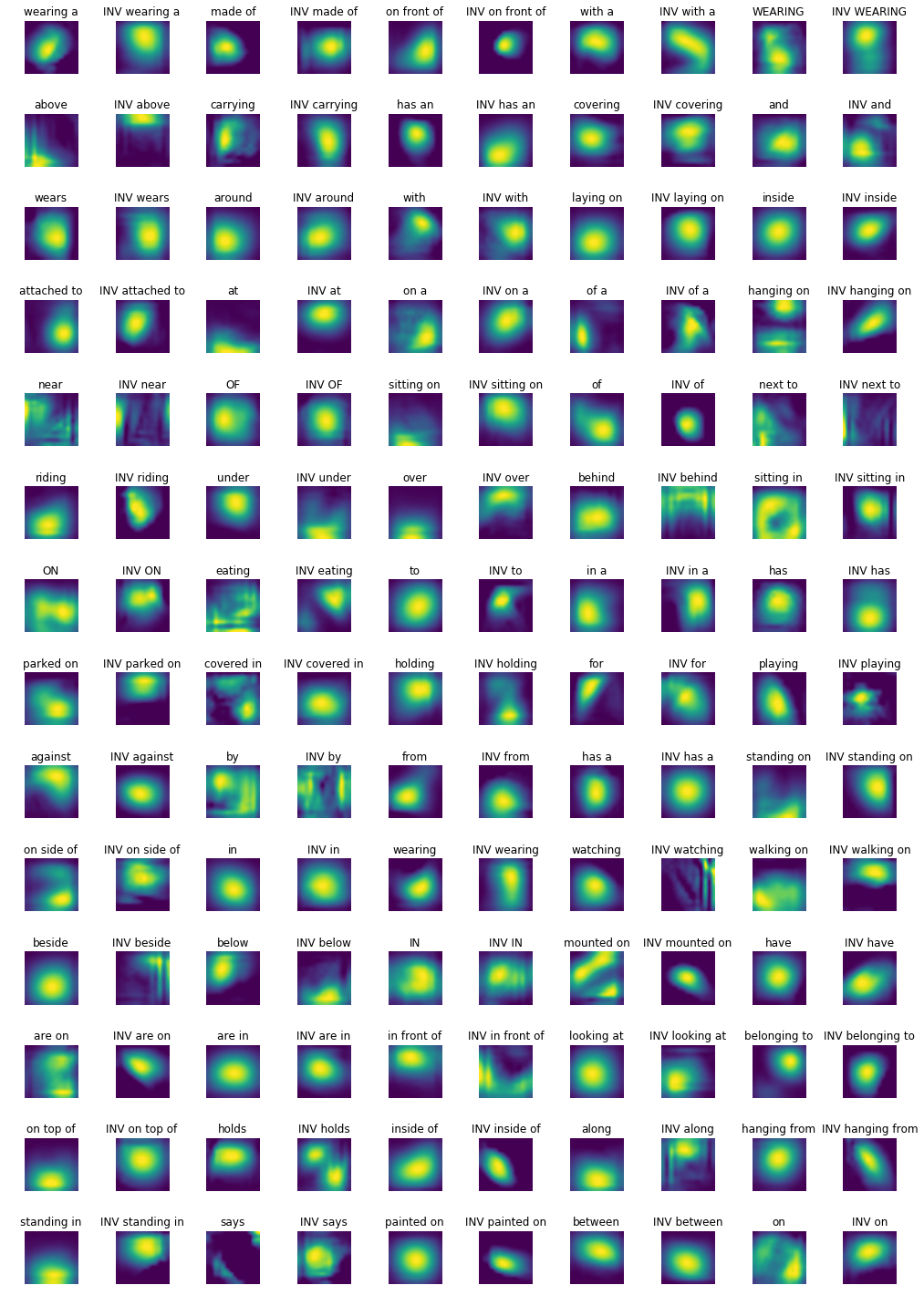}
    \caption{Learnt predicate shifts from the Visual Genome dataset.}
    \label{fig:visualgenome_predicates}
\end{figure*}

\begin{figure*}[t]
    \centering
    \includegraphics[width=0.8\linewidth]{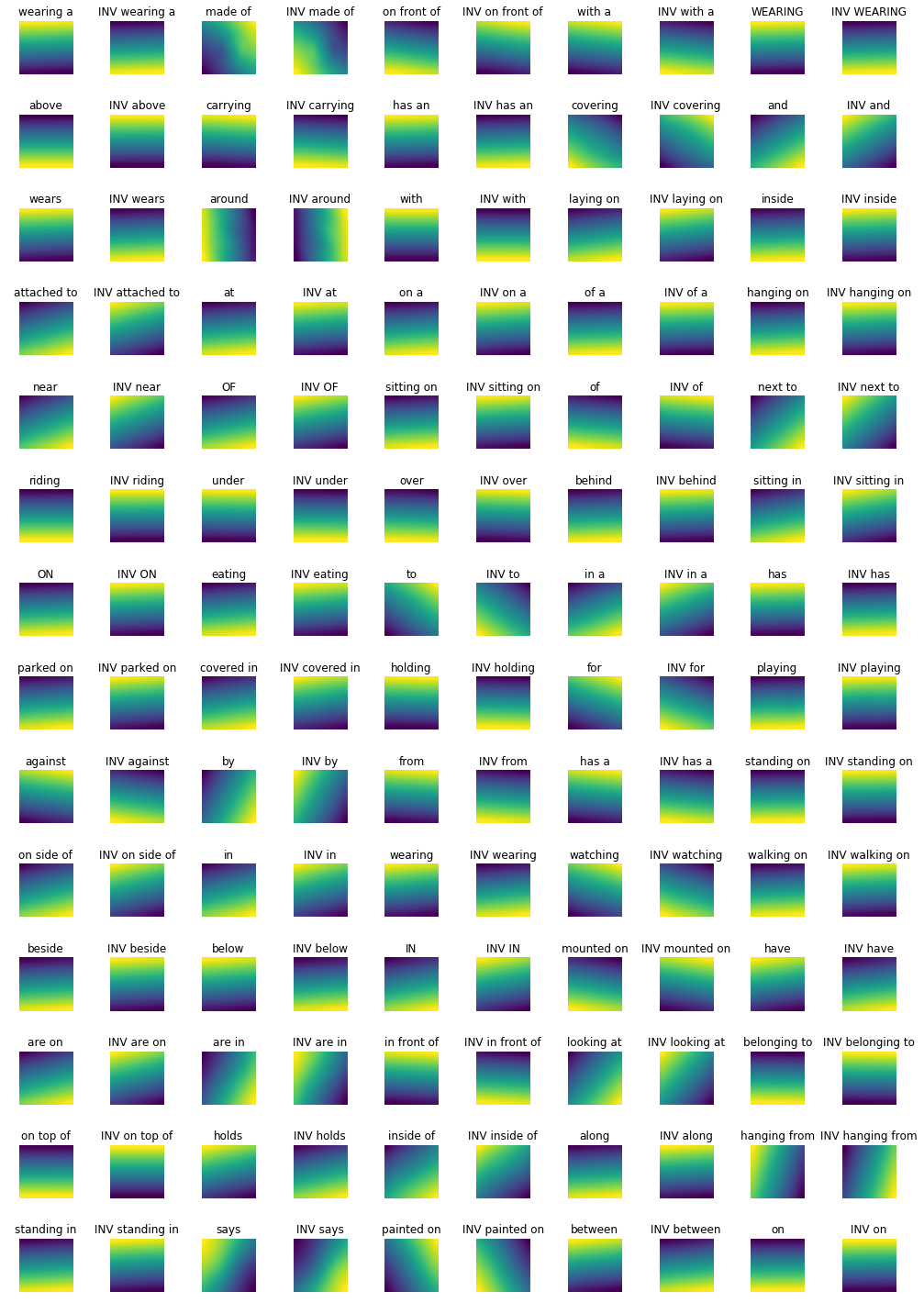}
    \caption{Spatial shifts calculated from the Visual Genome dataset. These shifts were used for the \textbf{spatial shift} baseline model.}
    \label{fig:visualgenome_gt_predicates}
\end{figure*}
\fi

{\small
\bibliographystyle{ieee}
\bibliography{egbib}
}
\end{document}